\pdfoutput=1

\documentclass[11pt]{article}

\usepackage{emnlp2021}

\usepackage{times}
\usepackage{latexsym}
\usepackage{amssymb}
\usepackage{graphicx}
\usepackage{booktabs}
\usepackage{subfigure}

\usepackage[T1]{fontenc}

\usepackage[utf8]{inputenc}

\usepackage{microtype}

\title{Blindness to Modality Helps Entailment Graph Mining}

\author{Liane Guillou$^{*\dag}$, Sander Bijl de Vroe$^{*\dag}$, Mark Johnson$^{\ddag}$, Mark Steedman$^{\dag}$ \\
  $^{\dag}$University of Edinburgh, $^{\ddag}$Oracle Digital Assistant \\
  \texttt{\{sbdv, liane.guillou\}@ed.ac.uk} \\
  \texttt{mark.mj.johnson@oracle.com, steedman@inf.ed.ac.uk} }

\begin{document}
\maketitle

\begingroup\renewcommand\thefootnote{*}
\footnotetext{Equal contribution}
\endgroup

\begin{abstract}
Understanding linguistic modality is widely seen as important for downstream tasks such as Question Answering and Knowledge Graph Population. Entailment Graph learning might also be expected to benefit from attention to modality. We build Entailment Graphs using a news corpus filtered with a modality parser, and show that stripping modal modifiers from predicates in fact increases performance. This suggests that for some tasks, the pragmatics of modal modification of predicates allows them to contribute as evidence of entailment.
\end{abstract}

\section{Introduction}

The ability to recognise textual entailment and paraphrase is crucial in many downstream tasks, including Open Domain Question Answering from text. For example, if we pose the question ``Did Joe Biden run for President?'' and the text states that ``Joe Biden was elected President'', producing the correct answer (Yes) necessitates understanding that \textit{being elected President} entails \textit{running for President}\footnote{Assuming a democratic election. We use the typical definition of the premise \textit{most likely} entailing the hypothesis \citep{dagan2006pascal}}.

Entailment Graphs, constructed via unsupervised learning techniques over large text corpora, provide a solution to this problem. They consist of nodes representing predicates and directed edges representing entailment relations between them. Given the importance of detecting uncertainty for other downstream NLP tasks such as Information Extraction \cite{karttunen2005veridicity, farkas2010conll}, Information Retrieval \cite{vincze2014uncertainty}, machine reading \cite{morante2012}, and Question Answering \cite{jean2016uncertainty} one might expect that it would also be useful in learning Entailment Graphs. That is, they would be more reliable if learned from data in which predications are \textit{asserted} as actually happening, rather than data with \textit{uncertain} predications under scope of various types of modality. We investigate whether this is the case.

The Entailment Graph-learning algorithm depends on descriptions of eventualities in the news, observing directional co-occurrences of typed predicates and their arguments. For example, we expect to observe all the arguments of \textit{being president}, such as \textit{Biden} and \textit{Obama}, also to be encountered in a sufficiently large multiply-sourced body of text as arguments of \textit{running for president}, but not the other way around (\textit{Hillary Clinton} will \textit{run} but not \textit{be president}). However, if all the reports of \textit{Clinton \textbf{might} be president} are extracted as \textit{be\_president(Clinton)}, one might expect the learning signal to be confusing to the algorithm.

We use the method of \citet{hosseini2018} combined with a modality parser \citep{bijldevroe2021modality} to construct typed Entailment Graphs from raw text corpora under two different settings. Modality-aware: modal predications are removed from the data entirely, and modality-unaware: the model learns from both asserted and modal predications. Our contributions are 1) a comparison of Entailment Graphs learned from modal and non-modal data, showing (counterintuitively) that ignoring modal distinctions in fact improves Entailment Graph-learning, and 2) insights as to whether this effect applies uniformly across different sub-domains. 

\section{Background}

\begin{table}
\small
\centering
\begin{tabular}{@{}ll@{}}
    \hline
    \textbf{Category} & \textbf{Example}\\
    \hline
    $\varnothing$ & Protesters attacked the police \\
    Modal operator & Protesters \textbf{may} have attacked the police \\
    Conditional & \textbf{If} protesters attack the police... \\
    Counterfactual & \textbf{Had} protesters attacked the police...\\
    Propositional & Journalists \textbf{said} that \\
    {    }attitude & protesters attacked the police \\
    \hline
\end{tabular}
\caption{Modality categories}
\label{table:modality_categories}
\end{table}

Entailment rules specify directional inferences between linguistic predicates \citep{szpektor2008},
and can be stored in an Entailment Graph, whose global structural properties can be used to learn more accurately \citep{berant2011, berant2015}. They are defined as a directed graph $\mathcal{G}=\{N,E\}$, in which the nodes $N$ are typed predicates and edges $E$ represent the entailment relation. The lexical entailment knowledge stored within them is useful for Question Answering \citep{mckenna2021}, as well as other tasks such as email categorisation \citep{eichler2014analysis}, relation extraction \citep{eichler-etal-2017-generating} and link prediction \citep{hosseini2019}.

A subgraph containing predicates of a type-pair (e.g. PERSON-LOCATION) can be learned in an unsupervised way from collections of multiply-sourced text. A vector of argument-pair counts for every predicate is first machine read from the corpus. Typically, relation extraction systems used for reading these corpora ignore modal modifiers, possibly introducing noise in the graph. Next, a (directed) similarity score (e.g. DIRT \citep{lin2001}, Weed's score \citep{weeds2003} or BInc \citep{szpektor2008}) is computed between the vectors, producing a local entailment score between each predicate pair. Then a globalisation process such as the soft constraints algorithm of \citet{hosseini2018}, which transfers information both within and between type-pair subgraphs, can be used to refine these local scores. When using the graph in practice, all edges with a score above a chosen threshold can be considered an entailment.

    There are various semantic phenomena a speaker can use to mark veridicality (see Table~\ref{table:modality_categories}). \textit{Modal operators}, e.g. \textit{probably, might, should, need to}, allow the user to indicate their attitude beyond the propositional content of a phrase, and often don't entail that the eventuality occurs \citep{kratzer2012modals}. The same holds for predications under scope of \textit{conditionals} and \textit{counterfactuals} \citep{dancygier1999conditionals, lewis1973counterfactuals}. \textit{Propositional attitude}, indicated by verbs such as \textit{say, imagine or want}, allows the speaker to attribute thoughts regarding some possible eventuality to a source \citep{mckay2000propositional}. 

These phenomena have been investigated for various NLP tasks, including uncertainty detection \citep{vincze2014uncertainty}, hedge detection \citep{medlock2007weakly} and modality annotation \citep{sauri2006annotating}. Capturing this information is valuable to tasks such as Information Extraction, Question Answering and Knowledge Base Population \citep{karttunen2005veridicity,morante2012annotating}.

Early approaches to detecting modality focused on lexicon design \citep{szarvas2008hedge, kilicoglu2008recognizing, baker2010}, with later approaches using machine learning over annotated corpora \citep{morante2009learning, rei2010combining, jean2016uncertainty, adel2016exploring}. Recently, \citet{bijldevroe2021modality} designed a parser similar to that by \citet{baker2010}, to cover a wider range of phenomena, including conditionality and propositional attitude. While modality annotation is clearly useful for recognising entailment from a given text \citep{snow2006effectively, de2006learning}, to our knowledge no research has been conducted on its effect on learning Entailment Graphs.

\section{Methods}

We extend relation extraction to pay attention to modality, so that we can distinguish modal and non-modal relations in the Entailment Graph mining algorithm. This allows us to investigate the impact of modalised predicate data on the accuracy of learned entailment edges. 

We extract \textit{binary relations} of the form arg1-predicate-arg2 using \textsc{MoNTEE}, an open-domain modality-aware relation extraction system \cite{bijldevroe2021modality}. \textsc{MoNTEE} uses the RotatingCCG parser \cite{stanojevic2019} as the basis for extracting binary relations and a modality lexicon to identify modality triggers. A relation is tagged as modal (MOD), propositional attitude (ATT\_SAY, ATT\_THINK) or conditional (COND) if the CCG dependency graph contains a path between a relation node and a node matching an entry in the \textsc{MoNTEE} lexicon. Counterfactuals (COUNT) are tagged according to hand-crafted rules. Since we focus on uncertainty and not negation, lexical negation (LNEG) tagging is ignored.

In the modality-aware setting, we remove relations tagged by \textsc{MoNTEE} as any kind of modal (\{MOD, ATT\_SAY, ATT\_THINK, COUNT, COND\}). In local learning, learned entailment edges then have access only to non-modal evidence: eventualities that were asserted as actually happening. For example, the edge between \textit{win} and \textit{lose} should now be learned only from  non-modal descriptions such as \textit{A won today against B} or \textit{A has been defeated by B}, leaving out modal descriptions (\textit{A could beat B}). The local and globalisation parts of the algorithm are otherwise unchanged.

\begin{figure*}[t!]
\begin{minipage}[b]{0.48\linewidth}
\centering
\includegraphics[width=\textwidth]{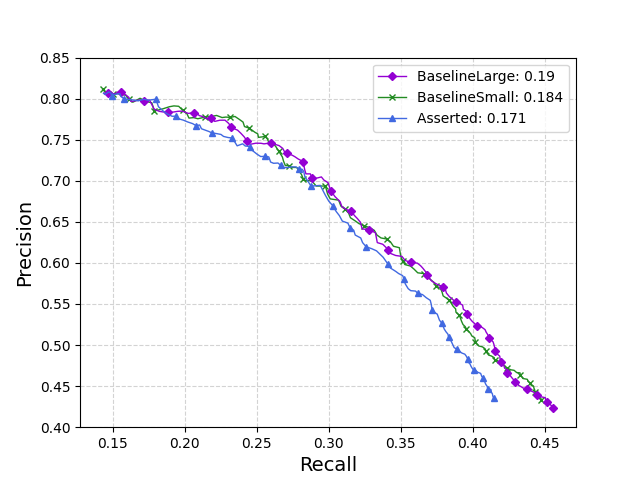}
\caption{Precision/recall on Levy/Holt dataset}
\label{fig:pr_rec_Levy_all_comb}
\end{minipage}
\hspace{0.5cm}
\begin{minipage}[b]{0.48\linewidth}
\centering
\includegraphics[width=\textwidth]{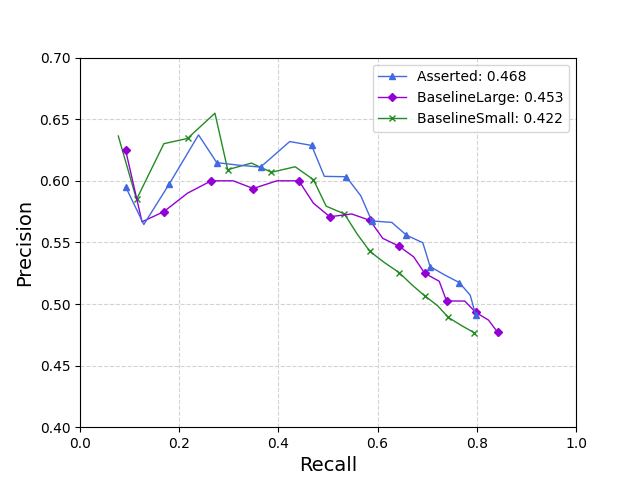}
\caption{Precision/recall on Sports Entailment Dataset}
\label{fig:pr_rec_Sports}
\end{minipage}
\end{figure*}

\section{Experimental Setup}

Using \textsc{MoNTEE}\footnote{https://gitlab.com/lianeg/montee}, we extract 40,669,812 binary relation triples from the NewsSpike corpus \citep{zhang2013}. Of these, 14.57\% are tagged; 10.04\% MOD, 3.51\% REP\_SAY, 0.38\% REP\_THINK, 0.61\% COND, and 0.03\% COUNT. We then construct three different datasets and build an Entailment Graph with each. The modality-unaware baseline, \textbf{BaselineLarge}, is trained on the complete set of relations with modality tags removed. This corresponds to the data and model in \citet{hosseini2018}. For the modality-aware \textbf{Asserted} graph, we extract only the set of 34,744,216 asserted relations ($\sim$85\% of the relations), i.e. all modal relations are excluded. To rule out effects of data size, we construct \textbf{BaselineSmall}, which is trained on a random sample of 85\% relations from the total set. Comparing Asserted to BaselineLarge shows us whether it is worth filtering out modal data, and comparing Asserted to BaselineSmall shows whether asserted data or mixed data (i.e. asserted and modal) is more effective for learning entailment relations.

We follow the example of \citet{hosseini2018} and construct typed graphs for all possible type pairs (e.g. PERSON-LOCATION). Relation arguments are typed by linking to a Named Entity Freebase identifier \citep{bollacker2008} using the AIDA-light linker \citep{nguyen2014}, and mapping these identifiers to a type in the FIGER hierarchy \citep{ling2012}. The typed relations become the input to the graph learning step of the Entailment Graph mining algorithm. Following previous research, we use the BInc similarity score \citep{szpektor2008} to compute entailment scores. We first construct local typed Entailment Graphs and then globalise the scores across graphs as in \citet{hosseini2018}.

We evaluate the Entailment Graphs on two datasets. The first is the \textit{Levy/Holt Entailment Dataset}, a set of 18,407 entailment pairs for the general domain \cite{levy2016, holt2018}. As our training method is unsupervised and we do not tune hyperparameters, we evaluate on the complete Levy/Holt dataset rather than the dev/test split. We also evaluate on the \textit{Sports Entailment Dataset} \cite{guillou2020incorporating}, focusing on the subset of 718 examples comprising entailments and pairs of match outcome predicates (e.g. \textit{win}, \textit{lose}, \textit{tie}, and their paraphrases) which are always non-entailments. This subset evaluates whether Entailment Graphs can recognise, for example, that win/lose $\rightarrow$ play but win $\nleftrightarrow$ lose (with similar patterns for other paraphrases of \textit{win, play} and \textit{lose}). We focus on the subgraph of ORGANISATIONs as all predicates are assumed to apply to sports teams. Both datasets use binary labels for each premise/hypothesis pair: entailment (1) and non-entailment (0).

We used the entGraph\footnote{https://github.com/mjhosseini/entGraph} code developed by \citet{hosseini2018} to construct each of the Entailment Graphs, and the corresponding evaluation scripts\footnote{https://github.com/mjhosseini/entgraph\_eval} to evaluate performance on the Levy/Holt dataset. Performance on the Sports Entailment Dataset\footnote{https://gitlab.com/lianeg/temporal-entailment-sports-dataset} is evaluated using scripts\footnote{https://gitlab.com/lianeg/sports-entailment-evaluation} developed for this paper. For details on hyperparameters and computational costs see Appendix~\ref{sec:appendix}.

\begin{table}[t]
\small
\centering
\begin{tabular}{lccc}
\toprule
& Levy/Holt & Levy/Holt & Sports \\
& all & directional & \\
\midrule
BaselineLarge & \textbf{0.190} & \textbf{0.163} & 0.453 \\
BaselineSmall & 0.184 & 0.157 & 0.422 \\
Asserted & 0.171 & 0.136 & \textbf{0.468} \\
\bottomrule
\end{tabular}
\caption{AUC scores}
\label{tab:auc_scores}
\end{table}

\begin{table}[t]
\small
\centering
\begin{tabular}{lcccc}
\toprule
& Nodes & Edges & \multicolumn{2}{c}{\% Levy preds found} \\
& & & all ex. & directional \\
\midrule
BaselineLarge & 334K & 72,7M & 63.06 & 70.29 \\
BaselineSmall & 277K & 58,4M & 61.13 & 69.29 \\
Asserted & 254K & 46,3M & 58.51 & 67.92 \\
\bottomrule
\end{tabular}
\caption{Graph size comparison and predicate coverage for Levy/Holt dataset (all examples) and its directional portion}
\label{tab:graph_size_comparison_Levy}
\end{table}

\begin{table}[t!]
\small
\centering
\begin{tabular}{lccc}
\toprule
& Nodes & Edges & \% Sports preds found \\
\midrule
BaselineLarge & 4,514 & 1.65M & 92.86 \\
BaselineSmall & 3,823 & 1.29M & 90.48 \\
Asserted & 3,682 & 1.09M & 88.10 \\
\bottomrule
\end{tabular}
\caption{ORGANISATIONs subgraph size comparison and predicate coverage for the Sports Entailment Dataset}
\label{tab:graph_size_comparison_Sports}
\end{table}

\section{Results}

Table~\ref{tab:auc_scores} contains area under the precision/recall curve (AUC) scores for Asserted, BaselineSmall, and BaselineLarge on the Levy/Holt dataset (\textbf{all} examples), the directional portion of the Levy/Holt dataset (2,414 examples), and the Sports Entailment dataset. The precision-recall curves for the Levy/Holt (all examples) and Sports Entailment datasets are displayed in Figures~\ref{fig:pr_rec_Levy_all_comb} and \ref{fig:pr_rec_Sports} respectively. Every point on the curve represents a different entailment score threshold (higher thresholds correspond to lower recall and vice versa). We follow the example of \citet{hosseini2018} and compute AUC for precision in the range [0.5, 1]. All three Entailment Graphs cover this range and predictions with precision higher than random are important for downstream applications.

On the Levy/Holt dataset (all examples), BaselineLarge performs best overall. The strong performance of BaselineLarge compared to Asserted is in itself surprising, and indicates that it is usually not beneficial to distinguish modality when building Entailment Graphs. This can be understood as a data size issue: filtering out data is harmful as it introduces sparsity, and modal data is useful enough to provide a learning signal.

More counterintuitive, however, is that even BaselineSmall, which controls for training dataset size, outperforms Asserted. To understand why, we measured the size of each graph in terms of the number of nodes (predicates) and edges (entailment relations) it contained, and the percentage of predicates in the Levy/Holt dataset that were present in the graph (see Table~\ref{tab:graph_size_comparison_Levy}). This revealed that BaselineSmall contained more of the predicates present in the Levy/Holt dataset, while also being larger in terms of both nodes and edges than Asserted. Thus, Asserted learns with more relations per predicate, while BaselineSmall has more predicate nodes overall. This may lead to the increase in recall that we see for the BaselineSmall graph.

Another explanation might be that this richer predicate coverage allows BaselineSmall to accurately correlate more of the common paraphrase examples in the Levy/Holt dataset. To this end we investigated the directional portion of the Levy/Holt dataset, which contains 2,414 examples of both the entailment pair and its reverse, where the entailment is true in one direction and false in the other. As noted by \citet{hosseini2018} this task is much harder than that represented by the original dataset. However, the baselines both outperform the Asserted graph on the directional entailment task. We also observe a similar pattern in the percentage of predicates covered (see last column in Table~\ref{tab:graph_size_comparison_Levy}). In general, we conclude that modal data is useful even for learning directional entailments.

Performance on the Sports Entailment dataset (Figure~\ref{fig:pr_rec_Sports}) reveals a different pattern. BaselineLarge outperforms BaselineSmall as expected, but Asserted performs best, despite lower coverage of the predicates in the Sports Entailment Dataset (see Table~\ref{tab:graph_size_comparison_Sports} for a size comparison of the ORGANISATIONs subgraph). This supports the suggestion by \citet{guillou2020incorporating} that excluding modal data may help to avoid learning entailments between disjunctive outcomes, i.e. that winning entails losing, which is not measured by the Levy/Holt dataset.

\section{Discussion \& Future Work}

Another appealing intuition for the usefulness of modal relations is that they might generally be expressed in text when the prior probability of the main predicate is already high. This would lead the distributions for the main predicates to be improved in spite of the uncertainty of the evidence. Additionally, if the probability of a premise is high enough to be worth mentioning, then in general that of its entailments will be too. However, this may not hold for the sports scenario because the outcomes are widely speculated upon despite being highly uncertain.

Indeed it is easy to find examples in the news corpus to support these intuitions. In the general domain we observe examples of eventualities initially being discussed with uncertainty, and later mentioned as asserted. An example of this is the acquisition of Dell by Michael Dell: on February 5th, 2013 we observe \textit{``... founder and CEO Michael Dell and investment firm Silver Lake Partners will buy Dell.''}, and subsequently, on February 6th, 2013 we read \textit{``So Michael Dell and a private equity group have bought Dell and taken it private.''}. We also observe the reverse scenario in the sports domain. For example, on January 10th, 2013 we observe \textit{``The popular opinion on this game seems to be Seattle beating Atlanta because..."}, while shortly afterwards we are informed that \textit{``Falcons come back to beat Seahawks"}. The latter is likely rather domain-specific, and we may expect to find a similar effect for other domains that share the disjunctive outcome property, for example elections, court cases and battles, where modals are used when speculating about potential and counterfactual outcomes.

We will explore ways to leverage this information and consider other sub-domains for which it is useful to retain or remove modal data. This may involve creating more domain-specific datasets. It is also worth investigating the effects of negation, which shares similar properties to modality, on learning Entailment Graphs. 

Relatedly, we could retain predicates under specific modal modifiers, as these correspond to different prior probabilities of eventualities, carrying a different epistemic commitment from the writer. Eventualities that happen ``undoubtedly" might be preferred over those that are ``unlikely", for instance, and the modality parser can output specific categories of modality, allowing us to choose the subsets that should be kept. 

Finally, we will experiment with learning Entailment Graphs with modal predicate nodes, by retaining modal relations with tags attached as input. Many of these entailments are trivial, because any entailment of a consequence can be reproduced under modal scope (if \textit{buy} $\rightarrow$ \textit{own}, then also \textit{\textit{MOD\_buy}} $\rightarrow$ \textit{\textit{MOD\_own}}). More notably, we might recover that following an entailment in the reverse direction can produce a modal entailment (e.g. if \textit{beat} $\rightarrow$ \textit{play}, then we know \textit{play} $\rightarrow$ \textit{\textit{MOD\_beat}}), and many preconditions will behave interestingly (e.g. \textit{beat} $\rightarrow$ \textit{play}, but also \textit{MOD\_beat} $\rightarrow$ \textit{play}). To evaluate this idea, we will design a dataset of modal entailments, drawing inspiration from previous research on veridicality in entailment datasets \citep{staliunaite2018learning}.

\section{Conclusion}
We have investigated the role of modally modified relations in Entailment Graph mining, and shown that, contrary to results from other tasks, uncertain predications actually constitute a valuable learning signal overall. Further analysis shows that there are specific predicate domains in which removing modal data is beneficial.

\section*{Acknowledgements}
This work was funded by the ERC H2020 Advanced Fellowship GA 742137 SEMANTAX and a grant from The University of Edinburgh and Huawei Technologies.

The authors would like to thank Mohammad Javad Hosseini and Nick McKenna for helpful discussions, and the reviewers for their valuable feedback.

\bibliography{emnlp2021}

\begin{thebibliography}{41}
\expandafter\ifx\csname natexlab\endcsname\relax\def\natexlab#1{#1}\fi

\bibitem[{Adel and Sch{\"u}tze(2017)}]{adel2016exploring}
Heike Adel and Hinrich Sch{\"u}tze. 2017.
\newblock \href {https://aclanthology.org/E17-1003} {Exploring different
  dimensions of attention for uncertainty detection}.
\newblock In \emph{Proceedings of the 15th Conference of the {E}uropean Chapter
  of the Association for Computational Linguistics: Volume 1, Long Papers},
  pages 22--34, Valencia, Spain. Association for Computational Linguistics.

\bibitem[{Baker et~al.(2010)Baker, Bloodgood, Dorr, Filardo, Levin, and
  Piatko}]{baker2010}
Kathryn Baker, Michael Bloodgood, Bonnie Dorr, Nathaniel~W. Filardo, Lori
  Levin, and Christine Piatko. 2010.
\newblock \href
  {http://www.lrec-conf.org/proceedings/lrec2010/pdf/446_Paper.pdf} {A modality
  lexicon and its use in automatic tagging}.
\newblock In \emph{Proceedings of the Seventh International Conference on
  Language Resources and Evaluation ({LREC}'10)}, Valletta, Malta. European
  Language Resources Association (ELRA).

\bibitem[{Berant et~al.(2015)Berant, Alon, Dagan, and Goldberger}]{berant2015}
Jonathan Berant, Noga Alon, Ido Dagan, and Jacob Goldberger. 2015.
\newblock \href {https://doi.org/10.1162/COLI_a_00220} {Efficient global
  learning of entailment graphs}.
\newblock \emph{Computational Linguistics}, 41(2):221--263.

\bibitem[{Berant et~al.(2011)Berant, Dagan, and Goldberger}]{berant2011}
Jonathan Berant, Ido Dagan, and Jacob Goldberger. 2011.
\newblock \href {https://www.aclweb.org/anthology/P11-1062} {Global learning of
  typed entailment rules}.
\newblock In \emph{Proceedings of the 49th Annual Meeting of the Association
  for Computational Linguistics: Human Language Technologies}, pages 610--619,
  Portland, Oregon, USA. Association for Computational Linguistics.

\bibitem[{Bijl~de Vroe et~al.(2021)Bijl~de Vroe, Guillou, Stanojevi{\'c},
  McKenna, and Steedman}]{bijldevroe2021modality}
Sander Bijl~de Vroe, Liane Guillou, Milo{\v{s}} Stanojevi{\'c}, Nick McKenna,
  and Mark Steedman. 2021.
\newblock \href {https://doi.org/10.18653/v1/2021.case-1.6} {Modality and
  negation in event extraction}.
\newblock In \emph{Proceedings of the 4th Workshop on Challenges and
  Applications of Automated Extraction of Socio-political Events from Text
  (CASE 2021)}, pages 31--42, Online. Association for Computational
  Linguistics.

\bibitem[{Bollacker et~al.(2008)Bollacker, Evans, Paritosh, Sturge, and
  Taylor}]{bollacker2008}
Kurt Bollacker, Colin Evans, Praveen Paritosh, Tim Sturge, and Jamie Taylor.
  2008.
\newblock \href {https://doi.org/10.1145/1376616.1376746} {Freebase: A
  collaboratively created graph database for structuring human knowledge}.
\newblock In \emph{Proceedings of the 2008 ACM SIGMOD International Conference
  on Management of Data}, SIGMOD '08, page 1247–1250, New York, NY, USA.
  Association for Computing Machinery.

\bibitem[{Dagan et~al.(2006)Dagan, Glickman, and Magnini}]{dagan2006pascal}
Ido Dagan, Oren Glickman, and Bernardo Magnini. 2006.
\newblock The pascal recognising textual entailment challenge.
\newblock In \emph{Machine learning challenges. evaluating predictive
  uncertainty, visual object classification, and recognising textual
  entailment}, pages 177--190. Springer.

\bibitem[{Dancygier(1998)}]{dancygier1999conditionals}
Barbara Dancygier. 1998.
\newblock \emph{Conditionals and Prediction: Time, Knowledge and Causation in
  Conditional Constructions}, volume~87 of \emph{Cambridge Studies in
  Linguistics}.
\newblock Cambridge University Press.

\bibitem[{De~Marneffe et~al.(2006)De~Marneffe, MacCartney, Grenager, Cer,
  Rafferty, and Manning}]{de2006learning}
Marie-Catherine De~Marneffe, Bill MacCartney, Trond Grenager, Daniel Cer, Anna
  Rafferty, and Christopher~D Manning. 2006.
\newblock Learning to distinguish valid textual entailments.
\newblock In \emph{Second Pascal RTE Challenge Workshop}, volume~62.

\bibitem[{Eichler et~al.(2014)Eichler, Gabryszak, and
  Neumann}]{eichler2014analysis}
Kathrin Eichler, Aleksandra Gabryszak, and G{\"u}nter Neumann. 2014.
\newblock An analysis of textual inference in german customer emails.
\newblock In \emph{Proceedings of the Third Joint Conference on Lexical and
  Computational Semantics (* SEM 2014)}, pages 69--74.

\bibitem[{Eichler et~al.(2017)Eichler, Xu, Uszkoreit, and
  Krause}]{eichler-etal-2017-generating}
Kathrin Eichler, Feiyu Xu, Hans Uszkoreit, and Sebastian Krause. 2017.
\newblock \href {https://doi.org/10.18653/v1/S17-1026} {Generating
  pattern-based entailment graphs for relation extraction}.
\newblock In \emph{Proceedings of the 6th Joint Conference on Lexical and
  Computational Semantics (*{SEM} 2017)}, pages 220--229, Vancouver, Canada.
  Association for Computational Linguistics.

\bibitem[{Farkas et~al.(2010)Farkas, Vincze, M{\'o}ra, Csirik, and
  Szarvas}]{farkas2010conll}
Rich{\'a}rd Farkas, Veronika Vincze, Gy{\"o}rgy M{\'o}ra, J{\'a}nos Csirik, and
  Gy{\"o}rgy Szarvas. 2010.
\newblock The conll-2010 shared task: learning to detect hedges and their scope
  in natural language text.
\newblock In \emph{Proceedings of the fourteenth conference on computational
  natural language learning--Shared task}, pages 1--12.

\bibitem[{Guillou et~al.(2020)Guillou, Bijl~de Vroe, Hosseini, Johnson, and
  Steedman}]{guillou2020incorporating}
Liane Guillou, Sander Bijl~de Vroe, Mohammad~Javad Hosseini, Mark Johnson, and
  Mark Steedman. 2020.
\newblock Incorporating temporal information in entailment graph mining.
\newblock In \emph{Proceedings of the Graph-based Methods for Natural Language
  Processing (TextGraphs)}, pages 60--71.

\bibitem[{Holt(2018)}]{holt2018}
Xavier Holt. 2018.
\newblock Probabilistic models of relational implication.
\newblock Master's thesis, Macquarie University.

\bibitem[{Hosseini et~al.(2018)Hosseini, Chambers, Reddy, Holt, Cohen, Johnson,
  and Steedman}]{hosseini2018}
Mohammad~Javad Hosseini, Nathanael Chambers, Siva Reddy, Xavier~R. Holt,
  Shay~B. Cohen, Mark Johnson, and Mark Steedman. 2018.
\newblock \href {https://doi.org/10.1162/tacl_a_00250} {Learning typed
  entailment graphs with global soft constraints}.
\newblock \emph{Transactions of the Association for Computational Linguistics},
  6:703--717.

\bibitem[{Hosseini et~al.(2019)Hosseini, Cohen, Johnson, and
  Steedman}]{hosseini2019}
Mohammad~Javad Hosseini, Shay~B. Cohen, Mark Johnson, and Mark Steedman. 2019.
\newblock \href {https://doi.org/10.18653/v1/P19-1468} {Duality of link
  prediction and entailment graph induction}.
\newblock In \emph{Proceedings of the 57th Annual Meeting of the Association
  for Computational Linguistics}, pages 4736--4746, Florence, Italy.
  Association for Computational Linguistics.

\bibitem[{Jean et~al.(2016)Jean, Harispe, Ranwez, Bellot, and
  Montmain}]{jean2016uncertainty}
Pierre-Antoine Jean, S{\'e}bastien Harispe, Sylvie Ranwez, Patrice Bellot, and
  Jacky Montmain. 2016.
\newblock Uncertainty detection in natural language: A probabilistic model.
\newblock In \emph{Proceedings of the 6th International Conference on Web
  Intelligence, Mining and Semantics}, pages 1--10.

\bibitem[{Karttunen and Zaenen(2005)}]{karttunen2005veridicity}
Lauri Karttunen and Annie Zaenen. 2005.
\newblock Veridicity.
\newblock In \emph{Dagstuhl Seminar Proceedings}. Schloss
  Dagstuhl-Leibniz-Zentrum f{\"u}r Informatik.

\bibitem[{Kilicoglu and Bergler(2008)}]{kilicoglu2008recognizing}
Halil Kilicoglu and Sabine Bergler. 2008.
\newblock Recognizing speculative language in biomedical research articles: a
  linguistically motivated perspective.
\newblock \emph{BMC bioinformatics}, 9(11):1--10.

\bibitem[{Kratzer(2012)}]{kratzer2012modals}
Angelika Kratzer. 2012.
\newblock \emph{Modals and conditionals: New and revised perspectives},
  volume~36.
\newblock Oxford University Press.

\bibitem[{Levy and Dagan(2016)}]{levy2016}
Omer Levy and Ido Dagan. 2016.
\newblock \href {https://doi.org/10.18653/v1/P16-2041} {Annotating relation
  inference in context via question answering}.
\newblock In \emph{Proceedings of the 54th Annual Meeting of the Association
  for Computational Linguistics (Volume 2: Short Papers)}, pages 249--255,
  Berlin, Germany. Association for Computational Linguistics.

\bibitem[{Lewis(1973)}]{lewis1973counterfactuals}
David Lewis. 1973.
\newblock Counterfactuals and comparative possibility.
\newblock \emph{Journal of Philosophical Logic}, pages 418--446.

\bibitem[{Lin and Pantel(2001)}]{lin2001}
Dekang Lin and Patrick Pantel. 2001.
\newblock \href {https://doi.org/10.1145/502512.502559} {{DIRT}: {D}iscovery of
  {I}nference {R}ules from {T}ext}.
\newblock In \emph{Proceedings of the Seventh ACM SIGKDD International
  Conference on Knowledge Discovery and Data Mining (KDD'01)}, pages 323--328,
  New York, NY, USA. ACM Press.

\bibitem[{Ling and Weld(2012)}]{ling2012}
Xiao Ling and Daniel~S. Weld. 2012.
\newblock Fine-grained entity recognition.
\newblock In \emph{Proceedings of the Twenty-Sixth AAAI Conference on
  Artificial Intelligence}, AAAI'12, page 94–100. AAAI Press.

\bibitem[{McKenna et~al.(2021)McKenna, Guillou, Hosseini, de~Vroe, and
  Steedman}]{mckenna2021}
Nick McKenna, Liane Guillou, Mohammad~Javad Hosseini, Sander~Bijl de~Vroe, and
  Mark Steedman. 2021.
\newblock Multivalent entailment graphs for question answering.
\newblock \emph{arXiv preprint arXiv:2104.07846}.

\bibitem[{Medlock and Briscoe(2007)}]{medlock2007weakly}
Ben Medlock and Ted Briscoe. 2007.
\newblock Weakly supervised learning for hedge classification in scientific
  literature.
\newblock In \emph{Proceedings of the 45th annual meeting of the association of
  computational linguistics}, pages 992--999.

\bibitem[{Morante and Daelemans(2009)}]{morante2009learning}
Roser Morante and Walter Daelemans. 2009.
\newblock Learning the scope of hedge cues in biomedical texts.
\newblock In \emph{Proceedings of the BioNLP 2009 workshop}, pages 28--36.

\bibitem[{Morante and Daelemans(2012{\natexlab{a}})}]{morante2012}
Roser Morante and Walter Daelemans. 2012{\natexlab{a}}.
\newblock \href
  {http://ceur-ws.org/Vol-1178/CLEF2012wn-QA4MRE-MoranteEt2012b.pdf}
  {Annotating modality and negation for a machine reading evaluation}.
\newblock In \emph{{CLEF} 2012 Evaluation Labs and Workshop, Online Working
  Notes, Rome, Italy, September 17-20, 2012}, volume 1178 of \emph{{CEUR}
  Workshop Proceedings}. CEUR-WS.org.

\bibitem[{Morante and Daelemans(2012{\natexlab{b}})}]{morante2012annotating}
Roser Morante and Walter Daelemans. 2012{\natexlab{b}}.
\newblock Annotating modality and negation for a machine reading evaluation.
\newblock In \emph{CLEF (Online Working Notes/Labs/Workshop)}, pages 17--20.

\bibitem[{Nelson(2019)}]{mckay2000propositional}
Michael Nelson. 2019.
\newblock {Propositional Attitude Reports}.
\newblock In Edward~N. Zalta, editor, \emph{The {Stanford} Encyclopedia of
  Philosophy}, {S}pring 2019 edition. Metaphysics Research Lab, Stanford
  University.

\bibitem[{Nguyen et~al.(2014)Nguyen, Hoffart, Theobald, and
  Weikum}]{nguyen2014}
Dat~Ba Nguyen, Johannes Hoffart, Martin Theobald, and Gerhard Weikum. 2014.
\newblock Aida-light: High-throughput named-entity disambiguation.
\newblock \emph{Workshop on Linked Data on the Web}, 1184:1--10.

\bibitem[{Rei and Briscoe(2010)}]{rei2010combining}
Marek Rei and Ted Briscoe. 2010.
\newblock Combining manual rules and supervised learning for hedge cue and
  scope detection.
\newblock In \emph{Proceedings of the Fourteenth Conference on Computational
  Natural Language Learning--Shared Task}, pages 56--63.

\bibitem[{Saur{\i} et~al.(2006)Saur{\i}, Verhagen, and
  Pustejovsky}]{sauri2006annotating}
Roser Saur{\i}, Marc Verhagen, and James Pustejovsky. 2006.
\newblock Annotating and recognizing event modality in text.
\newblock In \emph{Proceedings of 19th International FLAIRS Conference}.

\bibitem[{Snow et~al.(2006)Snow, Vanderwende, and
  Menezes}]{snow2006effectively}
Rion Snow, Lucy Vanderwende, and Arul Menezes. 2006.
\newblock Effectively using syntax for recognizing false entailment.
\newblock In \emph{Proceedings of the Main Conference on Human Language
  Technology Conference of the North American Chapter of the Association of
  Computational Linguistics}, HLT-NAACL '06, page 33–40. Association for
  Computational Linguistics.

\bibitem[{Stali{\=u}nait{\.e}(2018)}]{staliunaite2018learning}
Ieva~R. Stali{\=u}nait{\.e}. 2018.
\newblock Learning about non-veridicality in textual entailment.
\newblock Master's thesis, Utrecht University.

\bibitem[{Stanojevi{\'c} and Steedman(2019)}]{stanojevic2019}
Milo{\v{s}} Stanojevi{\'c} and Mark Steedman. 2019.
\newblock \href {https://doi.org/10.18653/v1/N19-1020} {{CCG} parsing algorithm
  with incremental tree rotation}.
\newblock In \emph{Proceedings of the 2019 Conference of the North {A}merican
  Chapter of the Association for Computational Linguistics: Human Language
  Technologies, Volume 1 (Long and Short Papers)}, pages 228--239, Minneapolis,
  Minnesota. Association for Computational Linguistics.

\bibitem[{Szarvas(2008)}]{szarvas2008hedge}
Gy{\"o}rgy Szarvas. 2008.
\newblock Hedge classification in biomedical texts with a weakly supervised
  selection of keywords.
\newblock In \emph{Proceedings of {ACL}-08: HLT}, pages 281--289.

\bibitem[{Szpektor and Dagan(2008)}]{szpektor2008}
Idan Szpektor and Ido Dagan. 2008.
\newblock \href {https://www.aclweb.org/anthology/C08-1107} {Learning
  entailment rules for unary templates}.
\newblock In \emph{Proceedings of the 22nd International Conference on
  Computational Linguistics (Coling 2008)}, pages 849--856, Manchester, UK.
  Coling 2008 Organizing Committee.

\bibitem[{Vincze(2014)}]{vincze2014uncertainty}
Veronika Vincze. 2014.
\newblock \emph{Uncertainty detection in natural language texts}.
\newblock Ph.D. thesis, University of Szeged.

\bibitem[{Weeds and Weir(2003)}]{weeds2003}
Julie Weeds and David Weir. 2003.
\newblock \href {https://www.aclweb.org/anthology/W03-1011} {A general
  framework for distributional similarity}.
\newblock In \emph{Proceedings of the 2003 Conference on Empirical Methods in
  Natural Language Processing}, pages 81--88.

\bibitem[{Zhang and Weld(2013)}]{zhang2013}
Congle Zhang and Daniel~S Weld. 2013.
\newblock Harvesting parallel news streams to generate paraphrases of event
  relations.
\newblock In \emph{Proceedings of the 2013 Conference on Empirical Methods in
  Natural Language Processing}, pages 1776--1786.

\end{thebibliography}
\bibliographystyle{acl_natbib}

\appendix
\section{Experimental Settings / Requirements}
\label{sec:appendix}
When using MoNTEE to extract relations we used the default settings, with the exception of disabling unary relation extraction (writeUnaryRels=False) and restricting binary relations to those that include at least one named entity (acceptGGBinary=False). When using entGraph to construct Entailment Graphs we raised the threshold values for infrequent predicates (minPredForArgPair=4) and argument pairs (minArgPairForPred=4), and used the default values for all other parameters. 

All experiments were conducted on a single server with 330GB RAM, and two Intel Xeon E5-2697 v4 2.3GHz CPUs (each with 18 cores). The computational cost of training a single Entailment Graph is approximately one day for the local learning step, and eight hours for globalisation.

\end{document}